\documentclass{fcs}
\usepackage{bm}
\usepackage{adjustbox}
\usepackage{caption}
\usepackage{multirow}
\usepackage{algorithm}
\usepackage{algorithmic}
\usepackage{rotating}
\usepackage{epstopdf}

\newcommand{\bs}{\boldsymbol}

\newcommand{\argmax}{\operatornamewithlimits{argmax}}
\graphicspath{{image/}}

\volumn{ }
\doi{ }
\articletype{RESEARCH~ARTICLE}
\title{$\bm{}$\\[2mm] Extractive and Abstractive Sentence Labelling of Sentiment-bearing Topics}
\author{Mohamad Hardyman BARAWI$^{1}$, Chenghua LIN $^{2}$, Advaith SIDDHARTHAN$^{3}$, Yinbin LIU $^{4}$ }
\address{{1\quad Computing Science, University of Aberdeen, UK}
\\
{2\quad Department of Computer Science, University of Sheffield,  UK}
\\
{3\quad Knowledge Media Institute, The Open University, UK}
\\
{4\quad School of Management, Shanghai University, China}}

\begin{document}
\maketitle
\setcounter{page}{1}
\setlength{\baselineskip}{14pt}

%\linenumbers

\begin{abstract}

This paper tackles the problem of automatically labelling sentiment-bearing topics with descriptive sentence labels. We propose two approaches to the problem, one extractive and the other abstractive. Both approaches rely on a novel mechanism to automatically learn the relevance of each sentence in a corpus to sentiment-bearing topics extracted from that corpus. The extractive approach uses a sentence ranking algorithm for label selection which for the first time jointly optimises topic--sentence relevance as well as aspect--sentiment co-coverage. The abstractive approach instead addresses aspect--sentiment co-coverage by using sentence fusion to generate a sentential label that includes relevant content from multiple sentences. To our knowledge, we are the first to study the problem of labelling sentiment-bearing topics. Our  experimental results on three real-world datasets show that both the extractive and abstractive approaches  outperform four strong baselines in terms of facilitating topic understanding and interpretation. In addition, when comparing extractive and abstractive labels,  our evaluation shows that our best performing abstractive method is able to provide more topic information coverage in fewer words, at the cost of generating less grammatical lables than the extractive method. We conclude that abstractive methods can effectively synthesise the rich information contained in sentiment-bearing topics.
\end{abstract}


\section{Introduction} \label{sec:intro}

Probabilistic topic models such as latent Dirichlet allocation (LDA) \cite{blei2003} capture the thematic properties of documents by modelling texts as a mixture of distributions over words,  known as topics. The words under each topic tend to co-occur together and consequently are thematically related to one another. These topics can therefore be used as a lens for exploring and understanding large archives of unstructured text. Since the introduction of LDA, many extensions have been proposed, an important one being the Joint Sentiment-Topic (JST) model that aims to mine and uncover rich opinion structures from opinionated documents~\cite{Lin2011}. This work has spurred subsequent research in developing variants of sentiment topic models for a range of opinion mining tasks such as aspect-based sentiment analysis~\cite{poria2016sentic},  contrastive opinion mining~\cite{ibeke2017unified}, and the analysis of sentiment and topic dynamics~\cite{si2013exploiting}.

\begin{table*}[th]
\small   \caption{Examples of LDA and JST topics with nouns (unformatted), adjectives(italics) verbs (bold).}
{\small 
\begin{adjustbox}{width=0.72\textwidth,center=\textwidth} 
\begin{tabular}{|l|}
\hline
{[}LDA{]}: computer models information data system network model \textit{parallel} methods software \\
{[}JST{]}: amazon order \textbf{return ship receive}  refund  \textit{damaged disappointed}  policy \textit{unhappy}                     \\ \hline
\end{tabular}
\end{adjustbox}
}
\label{tb:LDA-JST-example}
\end{table*}

JST is a hierarchical topic model which can detect sentiment and topic simultaneously from opinionated documents. The  hidden topics discovered, therefore,  are essentially sentiment-bearing topics resembling opinions. This is a key difference compared to the standard topics extracted by LDA which only express thematic information. We exemplify this difference using two topic examples  shown in Table~\ref{tb:LDA-JST-example}, where each topic is summarised using the ten topic words with the highest marginal probability. By examining the topic words, we see that the terms of the LDA topic are recognisably about the theme  \textit{``computer technology''}, whereas the terms of the JST topic capture  opinions relating to  \textit{``unsatisfactory online shopping experience''}. 

Although sentiment topic models have become an increasingly popular tool for exploring and understanding opinions from  text, existing models all share some noticeable drawbacks which can significantly limit their usefulness and effectiveness. First, applying sentiment topic models for exploratory purposes requires unfolding the meaning of the topics discovered,  which, so far, relies entirely on manual interpretation. However, it is generally difficult for a non-expert user  to understand a topic  based only on its multinomial distribution, especially when the user is not familiar with the source collection~\cite{mei2007automatic}.  
% In addition, even for experienced topic model users, manually interpreting the meaning of topics is slow and expensive.
The second limitation lies in these models' inability to facilitate accurate opinion and sentiment understanding, which is crucial for many applications. The issue stems from the fact that,  although the topic words of sentiment-bearing topics collectively express opinions, only a limited understanding of these opinions is possible by examining a list of topic words. Using the JST topic in Table~\ref{tb:LDA-JST-example} as an example, through  manual examination one can interpret that this topic expresses opinions relating to an \textit{``unsatisfactory online shopping experience''}, but it is impossible to  gain any deeper insight, e.g.  whether the sentiment  \textit{unhappy} is being expressed about the product being ordered  or about Amazon's policies. 

Remarkably, there is no existing work that has studied the problem of automatic labelling of sentiment-bearing topics. Substantial work on automatic topic labelling exists, but  has focused on labelling standard topics discovered by LDA to facilitate the interpretation of topics discovered.  These works note that  standard topics mainly express thematic information~\cite{aletras2015evaluating} and that most of the high ranking topic terms are nouns,  
%for which \textit{noun} is the dominant POS, 
and therefore aim to extract \emph{phrasal} labels that capture this thematic association to the topic terms. In contrast, the top terms of sentiment-bearing topics consist of a good mixture of nouns, verbs, and adjectives resembling opinions rather than themes~\cite{Lin2011,he2012online,he2013dynamic,li2013sentiment}. Existing topic labelling approaches  engineered for standard LDA topic are intrinsically unsuited to labelling sentiment-bearing topics as they provide no mechanism for modelling opinionated content. 

In this paper, we formally study the problem of automatically labelling sentiment-bearing topics extracted by the JST model~\cite{Lin2011}. In contrast to existing  approaches which generate topic labels in the form of either a single term~\cite{lau2010best}, a small set of terms (e.g. the top-$n$ topic words)~\cite{cano2014automatic}, or phrases~\cite{aletras2015evaluating}, we choose sentence as the label modality to facilitate interpretation of the opinions encoded in sentiment-bearing topics. Specifically, we propose two novel approaches for automatically generating  sentence labels that can facilitate  understanding and interpretation of multinomial sentiment-bearing topics extracted by JST, one extractive and the other abstractive. Both approaches rely on a novel mechanism to automatically learn the relevance of each sentence in a corpus to sentiment-bearing topics extracted from that corpus. The extractive approach uses a sentence ranking algorithm for label selection which for the first time jointly optimises this sentence relevance and aspect--sentiment co-coverage. The abstractive approach provides an alternate solution to aspect--sentiment co-coverage by using sentence fusion to generate a sentential label that includes relevant content from multiple sentences. %It is worth noting that the primary focus of our work is to automatically label sentiment-bearing topics, where the importance of topics with respect to the underlying documents is irrelevant to our task.
%The first approach generates extractive sentence labels in two stages: (i)  a mechanism is introduced to automatically learning the relevance between sentiment-bearing topics and the underlying sentences in a corpus; (ii) we then formulate a sentence  ranking algorithm for label selection which jointly considers topic-sentence relevance as well as aspect and sentiment co-coverage. The second approach  generates abstractive sentence labels following the sentence fusion paradigm. [\textbf{provide one or two more sentences to explain the fusion approach}]. 

We compare the effectiveness of our extractive topic labelling approaches against four strong extractive baselines, including two sentence label baselines and two  topic labelling systems which have been widely bench-marked in the literature~\cite{mei2007automatic,aletras2015evaluating}.  Experimental results  on three real-world datasets show that the labels generated by our approaches are more effective than all the baselines %and demonstrates the effectiveness of our sentence 
in terms of  facilitating sentiment-bearing topic understanding and interpretation. We then compare our extractive approach to two abstractive approaches and show that abstractive approaches can generate shorter labels with better coverage of the topic, but with some loss of grammaticality. To summarise, our contributions in this paper are three-fold: (i) we introduce a novel mechanism which can automatically learn the relevance to sentiment-bearing topics of the underlying sentences in a corpus; (ii) we design a sentence label selection criteria  which jointly considers relevance to the topic as well as aspect and sentiment co-coverage and show that it beats several strong baselines; (iii) we demonstrate the effectiveness of sentence fusion methods for generating abstractive labels which enhance topic coverage by including information from multiple sentences. 
%and (iii) we conduct an extensive set of experiments with both human evaluation and qualitative analysis on three real-world datasets. 

To our knowledge, we are the first to study the problem of labelling sentiment-bearing topics. We describe related work in Section~\ref{sec:related_work},  followed by our labelling method in Section~\ref{sec:methodology}, experimental design in Section~\ref{sec:setup}, results in Section~\ref{sec:results} and conclusions in Section~\ref{sec:conclusions}.

\section{Related Work} \label{sec:related_work}
%\textbf{[TODO:] we need to emphasis that our task focus on topic lablelling, rather than summarisation.}

Since its introduction, latent Dirichlet allocation (LDA) has become a popular tool for unsupervised analysis of text, providing both a predictive model of future text and a latent topic representation of the corpus. The latter property of topic models has enabled exploration and digestion of large text corpora through the extracted thematic structure represented as a multinomial distribution over words. 
%Topic models are used to extract topics in a corpus and are represented using the top-N words. 
Still, a major obstacle to applying LDA is the need to manually interpret the topics, which is generally difficult, especially for non-expert users. This has in turn motivated a number of approaches to automatically generating meaningful labels that facilitate topic understanding.

\subsection{Automatic Topic Labelling}

\textit{Automatically learning topic labels from data.}~~~In early work, Mei et al.~\cite{mei2007automatic} proposed generating topic labels using either bigrams or noun phrases extracted from a corpus and ranked the relevance of candidate labels to a given topic using KL divergence. Candidate labels were  scored using relevance functions which minimised the similarity distance between the candidate labels and the topic words. The top-ranked candidate label was then chosen as the topic's label. %Mei's approach showed that topic labels generated using ngrams were better in general when applied to different genre of texts. 
%The approach of \newcite{mei2007automatic} is somewhat computationally expensive because each topic's relevancy scores needs to be ranked with each of the top 1000 bigrams or top 1000 noun phrases.
Mao et al.~\cite{mao2012automatic}  labelled hierarchical topics by investigating the sibling and parent-child relations among the topics. Their approach followed a similar paradigm to \cite{mei2007automatic}, which ranked candidate labels by measuring the relevance of the labels to a reference corpus using Jensen-Shannon divergence. 
More recently, Cano et al.~\cite{cano2014automatic} proposed an approach to labeling topics based on multi-document summarisation. They measured the term relevance of documents to topics and generated topic label candidates based on the summarisation of a topic's relevant documents. 
%They then applied stop words removal and stemming to the news article headlines and used it as their Twitter topic labels.

\noindent \textit{Automatic topic labelling leveraging external sources.}~~~Another type of approach in automatic topic labelling leverages  external sources, e.g., Wikipedia or DBpedia~\cite{lau2010best,lau2011automatic,hulpus2013unsupervised,liu2014sherlock}. ~Lau et al.~\cite{lau2010best} proposed selecting the most representative term from a topic as its label by computing the similarity between each word and all others in the topic. Several sources of information were used to identify the best label, including pointwise mutual information scores, WordNet hypernymy relations, and distributional similarity. These features were combined in a re-ranking model. In a follow-up work, Lau et al.~\cite {lau2011automatic}  generated label candidates for a topic based on top-ranking topic terms and titles of Wikipedia articles. They then built a Support Vector Regression (SVR) model for ranking the label candidates. Other researchers~\cite{hulpus2013unsupervised} proposed using graph centrality measures to identify DBpedia concepts that are related to the topic words as their topic labels. More recently, Aletras et al.~\cite{aletras2015evaluating} proposed generating topic labels of multiple modalities, i.e., both  textual and images labels. To generate textual label candidates, they used the top-7 topic terms to search Wikipedia and Google to collect article titles which were subsequently chunk parsed. The image labels were generated  by first querying the top-$n$ topic terms to the Bing search engine. The top-$n$ candidate images retrieved were then ranked with PageRank and the image with the highest PageRank score  was selected as the topic label. More recently, \cite{aletras2017labeling} proposed to label topics using a pre-computed dependency-based word embeddings~\cite{levy2014dependency}, whereas 
\cite{bashar2017random} generated topic labels by utilising personalised domain ontologies.

%and use the output layer of VGG-net~\cite{simonyan2014very} pretrained on ImageNet~\cite{deng2009imagenet} to generate the caption of the images.
%Reviewer commented images labels=>image labels. What we meant here is using images as labels not an image with a label.

To summarise, the topic labelling approaches in the above-mentioned works were engineered for labelling standard LDA topics rather than sentiment-bearing topics, and mostly fall into one of the following linguistic modalities: a single term~\cite{lau2010best},  a small set of terms (e.g., the top-$n$ topic words)~\cite{lau2010best},  or phrases~\cite{lau2011automatic,aletras2015evaluating}. These linguistic modalities, while sufficient for expressing  facet subjects of standard topics, are inadequate to express complete thoughts or opinions due to their linguistic constraints, i.e. lacking a subject, a predicate or both. Our goal is to address this gap and study the problem of automatically labelling sentiment-bearing topics with more appropriate and descriptive sentence labels.

\subsection{Sentence Fusion}\label{ssec:sent_fusion}

Sentence fusion is a text-to-text generation technique that transforms overlapping information from a cluster of similar sentences into a single sentence. One of the most popular strategies for sentence fusion relies on merging the dependency trees of input sentences to produce a graph representation which will be linearised in a separate stage~\cite{barzilay2005,filippova2008sentence,elsner2011, cheung2014unsupervised}.  
%There are also works use the set theoretic concepts of union and intersection fusion~\cite{marsi2005}, which forgo the problem of identifying relevance and are thus less dependent on the context. These approaches extend fusion task that only uses the intersection of sentences~\cite{barzilay2005}, although both approaches point out the similar view that the best linearisation with a language model usually produces low rankings and not be able to deal with word order, agreement and sub-categorisation constraint.

Barzilay and McKeown~\cite{barzilay2005}  introduced a sentence fusion framework for multi-document summarisation. Their approach provides a means to capture the main information in a cluster of related sentences in comparison to the common method which just selects the sentence closest to the centroid. Approaches to sentence fusion typically use dependency graphs to produce an intermediary syntactic representation of the information in a cluster of sentences~\cite{barzilay2005,filippova2008sentence,elsner2011}. The graph is then linearised in different ways to generate fused sentences, which are then ranked using a language model or language-specific heuristics to filter out ill-formed sentences~\cite{thadani2013}. This type of approaches of fusing sentences mimic the strategies used by humans as reported in the analysis of human-generated summaries in~\cite{jing2000}. The sentence fusion task has since been widened to encompass other variants such as combining two sentences to produce a single sentence that either conveys the common information shared by the two sentences, or all information of sentences but without redundancy~\cite{marsi2005,krahmer2008query, elsner2011,nenkova2011automatic,cheung2014unsupervised}. 
More recent works~\cite{filippova2010multi,boudin2013,banerjee2015} on sentence fusion consider not only pairs of sentences but also larger sentence clusters using \textit{multi-sentence compression}, where the  compression is performed by selecting the high-scoring paths in a weighted bigram graph~\cite{filippova2010multi,boudin2013,banerjee2015}.

\section{Methodology} \label{sec:methodology}

The goal of our work is to develop  automatic approaches for labelling multinomial sentiment-bearing topics with descriptive sentence labels. We first briefly introduce the JST model used for extracting sentiment-bearing topics, and then proceed to describe the proposed approaches for sentence label generation. It should be noted  that our approaches do not have any specific dependencies on the JST model, and thus they are general enough to be directly applied to any other sentiment topic model variants that generate multinomial topics as output. 

\subsection{Preliminaries of the JST Model}
\begin{figure}[bt]
	 \centering 
%     \hspace{0.1in}
      %\subfloat[]{
    	%\label{fig:JST}
    	\includegraphics[width = 0.4\textwidth]{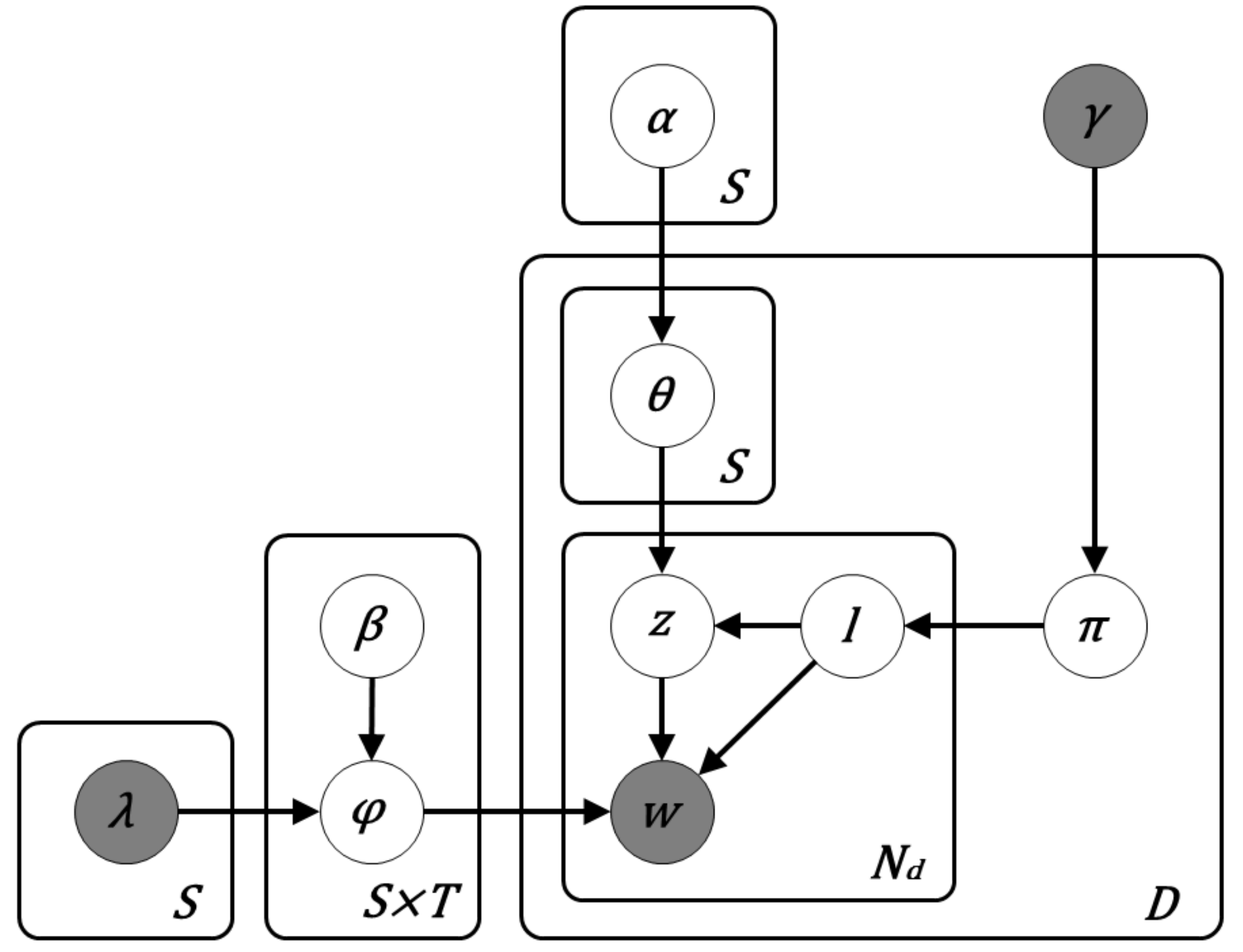}%}
        \hspace{0.2in}
        %\subfloat[]{
    	%\label{fig:generative-process}
    	%\includegraphics[width = 0.42\textwidth]{JST-generative.pdf}}
 	 \caption{JST graphical model.}\label{fig:JST}  %% label for entire figure
\end{figure}

The graphical model of JST is shown in Figure~\ref{fig:JST}, in which $D$ denotes a collection of documents, 
$N_{d}$ a sequence of words in document $d$, $S$  the number of  sentiment labels, $T$  the total number of topics, and $\{\alpha, \beta, \gamma\}$  the Dirichlet hyperparameters (cf.~\cite{Lin2011} for details).  %Fig.~\ref{fig:generative-process} summarises the complete procedure for generating a word $w_i$ in JST. 
The formal definition of the generative process in JST corresponding to the graphical model shown in Figure~\ref{fig:JST} is as follows. First, one draws a sentiment label $l$ from the per-document sentiment  proportion $\bs{\pi}_d$.
Following that, one draws a topic label $z$ from the per-document topic proportion $\bs{\theta}_{d,l}$ conditioned on sentiment label $l$. 
Finally, one draws a word from the per-corpus word distribution $\bs{\varphi}_{l,z}$ conditioned on both the sentiment label $l$ and topic label $z$.

\begin{itemize} 
\item For each sentiment label  $l \in\{1,...,S\}$ 
\begin{itemize}
\item  For each topic  $j \in\{1,...,T\}$  
\begin{itemize}
\item draw $\bs{\varphi}_{lj} \sim \text{Dir} (\lambda_{l} \cdot \beta^{T}_{lj} )$ 
\end{itemize}		
\end{itemize}  
\item For each document  $d \in\{1,...,D\},$	
\begin{itemize}
\item Choose a distribution $\pi_{d} \sim  \text{Dir}(\gamma)$ 
\item For each sentiment label  $l$ under document $d$ 
\begin{itemize} 
\item Choose a distribution $\theta_{d,l} \sim \text{Dir}(\alpha)$  
\end{itemize}
\item  For each word $w_{i}$ in document $d$ 
\begin{itemize} 
\item Choose a sentiment label $l_{j} \sim \text{Mult}(\pi_{d})$ 
\item  Choose a topic $z_{i} \sim \text{Mult}(\theta_{d,l_{i}})$  
\item  Choose a word $w_{i} \sim \text{Mult}(\varphi_{l_{i}z_{i}})$ 
\end{itemize}
\end{itemize}
\end{itemize}

%There are three hyperparameters in JST. The $\alpha$ parameter is interpreted as the prior observation counts for the number of times a topic $j$ is associated with a sentiment label $l$ sampled from a document. The $\beta$ parameter is interpreted as the number of times words sampled from a topic $j$ are associated with a sentiment label $l$ before the observation of any actual words.  The hyperparameter $\gamma$ is interpreted as the prior observation counts for the number of times a sentiment label $l$ is sampled from a document before any word is observed from the corpus. In our implementation, we used an asymmetric prior $\alpha$ and symmetric prior $\beta$ and $\theta$. Also, there are three sets of latent variables that we need to infer in JST, i.e., the per-document sentiment distribution $\pi$, the per-document sentiment label specific topic distribution $\theta$, and the per- corpus joint sentiment-topic word distribution $\varphi$.
% * <advaith@abdn.ac.uk> 2017-11-30T13:06:21.996Z:
% 
% The above paragraph was impossible to read. I have tried to make the sentences grammatical, but I cannot still understand what you are saying. Where does Theta suddenly come from in "In our implementation, we used an asymmetric prior $\alpha$ and symmetric prior $\beta$ and $\theta$"? Do you mean gamma? Are priors the same as hyperparameters?
% 
% ^ <linron84@hotmail.com> 2017-12-01T10:45:56.091Z.

The problem to be solved by the JST model is the posterior inference of the variables, which determine the hidden sentiment-bearing topic structures that can best explain the observed set of documents. Formally, a sentiment-bearing topic is represented by a multinomial distribution over words denoted as $p(w|l,z) \equiv \varphi_{w,l,z}$, where $w$, $l$, and $z$ represent the word, sentiment label and topic label indices, respectively. In particular, the approximated per-corpus sentiment-topic word distribution is 
%\noindent\resizebox{7.7cm}{!} {
%\begin{minipage}{0.5\textwidth}
\begin{equation}\label{eq:phi}
\varphi_{l,z,w} = \frac{N_{l,z,w}+\beta} {N_{l,z}+ V \beta}, 
\end{equation}
%\end{minipage}}
where $V$ is the corpus vocabulary size, $N_{l,z,w}$ is the number of times word $w$ appeared in topic $z$ with sentiment label $l$, $N_{l,z}$ is the number of times words are assigned to $z$ and  $l$. Recalling that $\varphi_{w,l,z}$ is a multinomial distribution, we have $\sum_{w \in V} \varphi_{w,l,z} = 1$.

\subsection{Modelling the Relevance between Sentiment-bearing Topics and Sentences} \label{sec:sentence_generate}

The original JST model can only learn topic--word and topic--document associations as it operates on bag-of-words features at the document-level, with the corpus sentence structure being ignored. Therefore, we propose a new computational mechanism that can uncover the relevance to a sentiment-bearing topic of the underlying sentences in the corpus. To achieve this, we first preserve the sentence structure information for each document during the corpus preprocessing step (see Section~\ref{sec:setup} for more details). Second, modelling topic-sentence relevance is essentially equivalent to calculating the probability of a sentence given a sentiment-bearing topic  $p(\text{sent}|l,z)$, i.e., the likelihood of a sentence (from the corpus) associating with a given sentiment-bearing topic. The posterior inference of JST, based on Gibbs sampling~\cite{steyvers2007probabilistic,Lin2011},  can recover the hidden sentiment label and topic label assignments for each word in the corpus. Such label-word assignment information provides a means to re-assembling the relevance between a word and a sentiment-bearing topic. By leveraging the sentence structure information and gathering the label assignment statistics for each word of a sentence, we can derive the probability of a sentence given a sentiment-bearing topic as 
%\noindent\resizebox{7.8cm}{!} {
%\begin{minipage}{0.5\textwidth}
\begin{align}\label{eq:sent-posterior}
p(\text{sent}|l,z) &= \frac{p(l,z|\text{sent}) \cdot p(\text{sent})} {p(l,z)} \nonumber \\ 
& \propto p(l,z|\text{sent}) \cdot p(\text{sent}),
\end{align}
%\vspace{0.01mm}
%\end{minipage}}
where 
\begin{equation}
p(l,z|\text{sent})= \frac{\sum_{l'_w=l,z'_w=z}^{w \in \text{sent} \vee} \varphi_{l',z',w}}{ \sum_{w \in \text{sent}} \bs{\varphi}_{l',z',w}}
\end{equation}
\begin{equation}
p(\text{sent}) = \sum_{l}\sum_{z}\prod_{w \in \text{sent}}\varphi_{l,z,w}.
\end{equation} 
Note that the per-corpus sentiment-topic word  distribution $\varphi_{w,l,z}$, defined in Eq.~\ref{eq:phi}, is obtained via the posterior inference using Gibbs sampling. Also $p(l,z)$ is discounted as it is a constant when comparing sentencial labels for the same sentiment-bearing topic. A summary of our  mechanism of calculating $p(\text{sent}|l,z)$ is given in Algorithm~ \ref{alg:JST}.

%\begin{tcolorbox}[blanker,float=tb, grow to left by=-0.7cm,grow to right by=-0.7cm]
%\begin{wrapfigure}{R}{0.4\textwidth}
%\begin{minipage}{0.5\textwidth}
\begin{algorithm}[tb] \small \centering
\caption{Procedures of calculating $p(\text{sent}|l,z)$.} \label{alg:JST}
\begin{algorithmic}[1]
\REQUIRE{A corpus $C$} \ENSURE{The probability of observing a sentence given a sentiment-bearing topic $p(\text{sent}|l,z)$.}
\STATE Perform automatic sentence segmentation on corpus $C$ to preserve the  sentence structure information. 
\STATE Train a JST model based on $C$ following the standard procedures and settings described in~\cite{Lin2011}.
\FOR{each sentiment label  $l \in\{1,...,S\}$}
	\FOR{each topic label $z \in\{1,...,K\}$}
		\FOR{each sent $\in C$}
			\STATE Calculate $p(\text{sent}|l,z)$ based on Eq.~\ref{eq:sent-posterior}
		\ENDFOR	
		\STATE Normalise the probability of $p(\text{sent}|l,z)$
		\STATE Sort  sentences based on the probability of  $p(\text{sent}|l,z)$ in a descending order. 		
     \ENDFOR
\ENDFOR
\end{algorithmic}
\end{algorithm}
%\end{minipage}
%\end{wrapfigure}
%\end{tcolorbox}

\subsection{Extractive labelling of sentiment-bearing topics}

%\subsubsection{Automatic candidate sentence label selection} \label{sec:sentence_selection}

Given a sentiment-bearing topic, one intuitive approach for selecting the most representative sentence label is to rank the sentences in the corpus according to the topic-sentence relevance probability $p(\text{sent}|l,z)$ derived in the previous section. 
One can  then select the sentence with the highest probability  as a label for the topic. We found that a high-degree of relevance, however, cannot be taken as the only criterion for  label selection as it might result in the selection of short sentences that  do not provide enough information about the topic or that cover either thematic or sentiment information alone.
%A high-degree of relevance, however,  cannot be taken as the only criterion for sentence selection  as it might risk selecting a subset of very redundant sentences. For instance,  a sentiment-bearing topic about \textit{movie} might cover several related, sentiment-coupled 
%s; e.g. \textit{director, film, casting}. If one aspect of the topic, e.g. \textit{director}, is particularly prevalent, this approach will select all sentences about this aspect, and ignore all other aspects. In addition, the sentiments of the topic aspects often exhibit different sentiment intensity; i.e. \textit{director} is a more prevalent aspect than \textit{casting} but the associated sentiment with the latter is much stronger than the former. 
To address these issues and to select the most representative sentence labels covering sentiment-coupled aspects which reveal opinions, it is important to consider the occurrences of both aspect and sentiment, and the balance between them. To this end, we designed a sentence label selection criterion which jointly considers relevance as well as aspect and sentiment co-coverage. This provides the basis for building an algorithm for ranking sentence labels in terms of how well they can describe the opinion encoded in a sentiment-bearing topic. 

%Such criteria provides a basis for building an ILP framework~\cite{mcdonald,gillick} for automated the sentence label selection.

%\subsubsection{Sentence label ranking}

Given a sentiment-bearing topic, we define the sentence scoring function below: 
\begin{equation}\label{eq:SMAC} 
L(s|t_{l,z}) =  \alpha \cdot Rel(s|t_{l,z}) + (1-\alpha) \cdot Cov(s|t_{l,z}). 
\end{equation}
Here, $Rel(s|t_{l,z})$ is the relevance score between a sentence $s$ and a given sentiment-bearing topic $t_{l,z}$ as described in Section 3.2. $Cov(s|t_{l,z})$, the aspect and sentiment co-coverage score, encodes two heuristics: (i) sentence labels covering either sentiment or aspects information alone will be significantly down-weighted; and (ii) labels which cover salient aspects of the topic coupled with sentiment will be given high weightage. Parameter $\alpha$  sets the relative contributions of relevance  and the co-coverage of aspect and sentiment, and was empirically set to 0.4. 
Let $i$ be a word from sentence $s$,  $a_{i}$  a binary variable that indicates whether word $i$ is an aspect word, $o_{i}$ a binary variable that indicates whether word $i$ is a sentiment word, $w_{l,z,i}$ the importance weight of word $i$ given topic $t_{l,z}$, and $t_{l,z,n}^{i}$ a binary variable which indicates the presence of word $i$ in the top-$n$ words of sentiment-bearing topic $t_{l,z}$. Formally, the aspect and sentiment co-coverage score is formulated as
\begin{equation}\label{eq:AScoverage}
Cov(s|t_{l,z}) = \frac{2 \cdot A(s|t_{l,z})\cdot S(s|t_{l,z})}{A(s|t_{l,z})+ S(s|t_{l,z})} 
\end{equation}
\begin{equation}
%A(s|t_{l,z}) = \sum_{w_{i} \in s } \varphi_{l,z,w_{i,a}} \cdot a_{s,i} \cdot t_{l,z,n}^{i}
A(s|t_{l,z}) = \sum_{i \in s } w_{l,z,i} \cdot a_{s,i} \cdot t_{l,z,n}^{i}
\end{equation}
\begin{equation}
S(s|t_{l,z}) = \sum_{i \in s} w_{l,z,i} \cdot o_{s,i} \cdot t_{l,z,n}^{i},
\end{equation}
%Similarly, $S(s) = \sum_{w_s \in s} \varphi_{l,z,w_a}$ xxx 
where the weight $w_{l,z,i}$ essentially equals to $\varphi_{l,z,i}$, i.e., the marginal probability of word $i$ given a sentiment-bearing topic $t_{l,z}$.  We identify whether word $i$ is an aspect word or a sentiment word as follows. In the preprocessing, we perform parts of speech (POS) tagging on the experimental datasets (detailed in Section~\ref{sec:setup}). 
%\footnote{\url{http://www.cs.pitt.edu/mpqa/}}
Words tagged as nouns are regarded as aspects words whereas sentiment words are the ones that  have appeared in the MPQA sentiment lexicon\footnote{http://www.cs.pitt.edu/mpqa/}. Recalling that a multinomial topic is  represented by its top-$n$ topic words with the highest marginal probability, we further constrain that word $i$ must also appear in the top-$n$ topic words of sentiment-bearing topic $t_{l,z}$. This ensures that a good sentence label should cover as many important sentiment coupled aspects of a topic as possible. In the case where a sentence contains either aspect or sentiment information alone (i.e., $A(s|t_{l,z})=0$ or $S(s|t_{l,z})=0$), the resulting aspect and sentiment co-coverage score would be zero,  thus down-weighting the corresponding sentence. Finally, a sentence label $\hat{s}$ for a sentiment-bearing topic $t_{l,z}$ can be obtained by: 
\begin{equation}\label{eq:object} 
\displaystyle \hat{s} = \argmax_s L(s|t_{l,z}). 
\end{equation}
% * <advaith@abdn.ac.uk> 2017-11-30T14:18:17.907Z:
% 
% provide the value on "n" used in the paper
% 
% ^.

\subsection{Abstractive topic labelling with multi-sentence compression} \label{sec:sentence_Fusion}

Our second type of approach to topic labelling employs sentence fusion techniques for generating abstractive labels. We hypothesise that abstractive sentence labels would be more succinct by removing redundant information of input sentences, while maintaining the essence of sentiment-bearing topics as much as possible. Specifically, we explore two well established sentence fusion algorithms~\cite{filippova2010multi,boudin2013} for the fusion task,
%namely PathGraph~\cite{filippova2010multi} and Keyphrase~\cite{boudin2013}. 
both of which search for an optimal weighting function in noisy graphs to identify readable and informative compressions.

\subsubsection{Word graph-based multi-sentence compression} \label{ssec:filipova}

\begin{figure*}[tb]
	 \centering 
%     \hspace{0.1in}
      %\subfloat[]{
    	%\label{fig:JST}
    	\includegraphics[width = 0.65\textwidth]{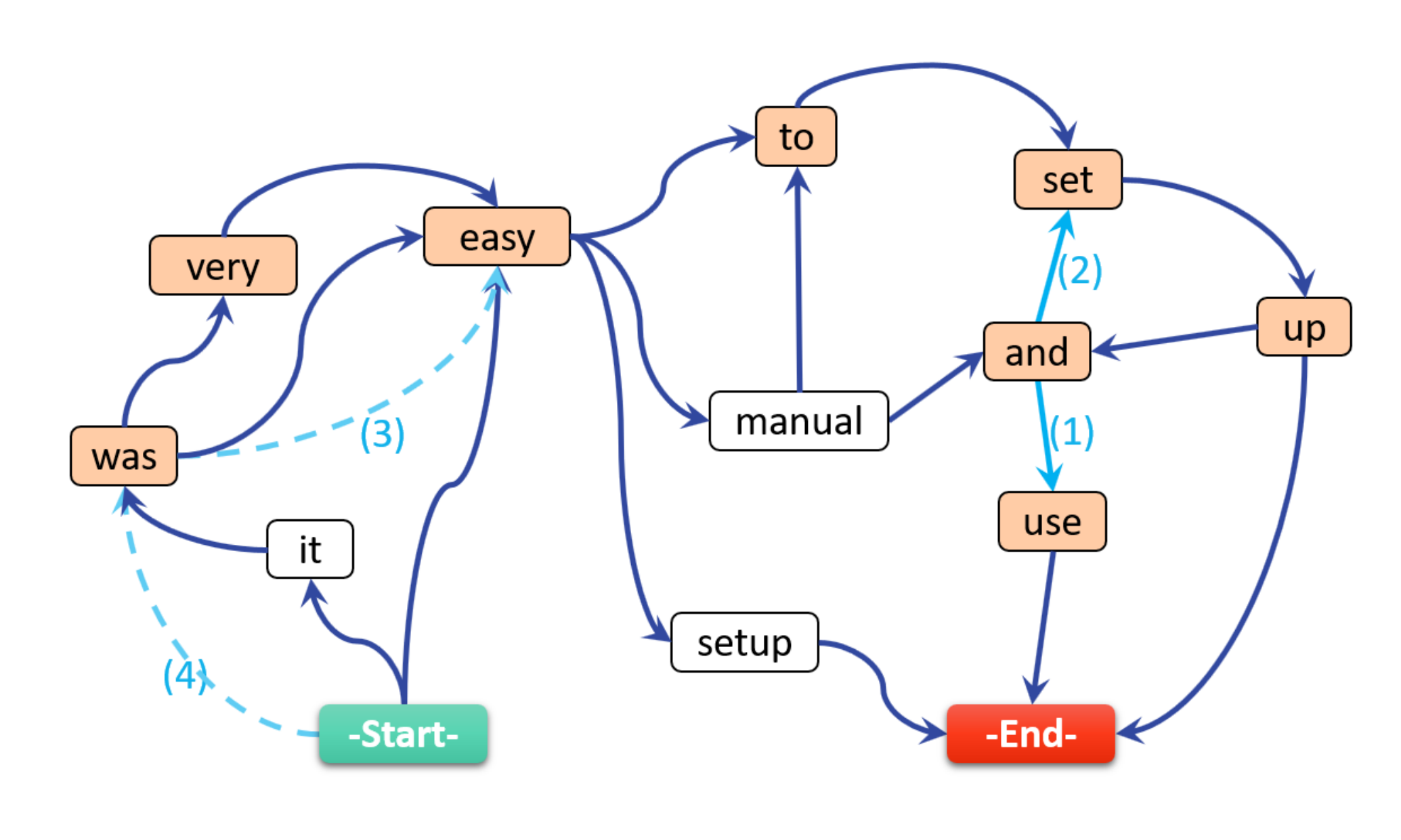} %}
        \hspace{0.2in}
        %\subfloat[]{
    	%\label{fig:generative-process}
    	%\includegraphics[width = 0.42\textwidth]{JST-generative.pdf}}
 	 \caption{Word graph constructed from four related sentences where a possible compression path is also given.}\label{fig:Word Graph}  %% label for entire figure
\end{figure*}

Our first sentence fusion approach (named \textbf{PathGraph}) extends the algorithm of Filippova~\cite{filippova2010multi} to produce well punctuated and informative compression. Taking the sentences extracted in Section 3.2 as input, we build a word graph from a set of related sentences  $S = \{s_{1}, s_{2}, ..., s_{n}\}$, by iteratively adding sentences to it.  Figure~\ref{fig:Word Graph} is an illustration of the word graph constructed from the four sentences below. For clarity, edge weights are omitted, and italicised fragments from the sentences are replaced with dots.
\begin{enumerate}
\item It was easy to set up and use.
\item Easy manual and set up.
\item It was \textit{a super} easy setup.
\item \textit{The D-Link DP-300U} was very easy to set up. 
\end{enumerate}

The word graph is constructed as follows. First, the first sentence is added to the graph represented by a string of word nodes plus the start and the end symbols, i.e.,  the \texttt{start} and \texttt{end} nodes in Figure~\ref{fig:Word Graph}. A word from the remaining sentences is mapped onto an existing node in the graph if they have the same lowercased word form and  part of speech (POS), in addition to that there is no word from the sentence that is already mapped onto the node. A new node is created if there is no suitable candidate node in the word graph. Words are added to the graph in the three steps: (i) non-stopwords for which no candidate exists in the graph or for which a non-ambiguous mapping is possible; (ii) non-stopwords for which there are either several possible candidates in the graph or which appear more than once in the sentence; and finally (iii) stopwords.

In the original algorithm of Filippova~\cite{filippova2010multi}, punctuation marks are excluded~\cite{filippova2010multi}. We add an additional step to generate well-punctuated compressions by adding punctuation marks in the graph following~\cite{boudin2013}, and select the candidate that has the same immediate context if the mapping is ambiguous.  Once all the words from a sentence are added to the graph,  words adjacent in the sentence are connected with directed edges, where the edge weights are calculated using the weighting function defined in Eq.~\ref{eq:edge_weight} and Eq.~\ref{eq:coherence_compression}.
%Next, the compressions are reranked by path length~\cite{filippova2010multi} with the K-shortest paths algorithm to find the fifty shortest paths scored in the function below:
\begin{equation}\label{eq:edge_weight} 
w(i,j)=\frac{\text{coherence}(i,j)} {\text{freq}(i) \times \text{freq}(j)}
\end{equation}
\begin{equation}\label{eq:coherence_compression} 
\text{coherence}(i,j) =\frac{\text{freq}(i) + \text{freq}(j)} { \sum_{s\in S}  \text{dist} (s,i,j)^{-1}}
\end{equation}
Here $\text{freq}(j)$ is the number of words mapped to the node $j$. The function $\text{dist}(s,i,j)$ refers to the distance between the offset positions of words $i$ and $j$ in sentence $s$. The purpose of this function is two-fold: (i) to generate a grammatical compression, it favours links between words which often occur in this order (see Eq.~\ref{eq:coherence_compression}); (ii) to generate an informative compression, it promotes paths passing via salient nodes. 

%However, this function only indicates how strong the association between two words is. To generate a compression that contains an informative compression, the weight of edges connecting salient nodes is decreased with respect to the frequency of the connected nodes as shown in Equation~\ref{eq:edge_weight}. 
% * <advaith@abdn.ac.uk> 2017-11-30T14:33:43.821Z:
% 
% I do not understand the above sentence at all.  Why do you want to reduce weight for frequent words? and why by multiplying frequencies? Please explain. Is this the formula used by Filippova? If so, say this, otherwise explain why we are doing something different
% 
% ^.
%An implementation of K-shortest paths algorithm is used to find the 50 shortest paths from start to end nodes in the graph. We added an additional constraint in the path selection to only select paths containing at least one noun and  topic word. Initially, the fusion approaches~\cite{filippova2010multi,boudin2013} only selects shorter paths which contain less than eight words or paths not containing a verb. Finally, the remaining paths are re-ranked by normalising the total path weight over its length.

%An implementation of K-shortest paths algorithm is used to find the 50 shortest paths from start to end nodes in the graph. 

To produce more informative sentences which maximise the range of topics they cover, we further make use of the top-15 topic words extracted by JST. The rationale behind is that topic words can capture the gist of documents, and thus can be used to better generate sentences that convey the gist of the set of related sentences. To do so, we added syntactic constraints in the path selection process to filter all the paths which do not contain any topic words (among the top-15 topic words) for the topic being labelled,  which do not pass a verb node and which are shorter than eight words. Finally, the remaining paths are re-ranked by normalising the total path weight  over its length.
\subsubsection{Word graph compression by keyphrases relevance}\label{ssec:boudin}

The second sentence fusion algorithm (named \textbf{Keyphrase}) extends the work of~\cite{boudin2013}, which is similar to GraphPath but differs in that it re-ranks the generated compressions according to the number relevant keyphrases a sentence compression contains. Specifically, keyphrases are extracted from the cluster of related sentences in two steps. First, a weighted graph is constructed from the set of related sentences, where nodes represent tuples consisting of lowercased words as well as their corresponding parts of speech. Two nodes are connected if their corresponding lexical units co-occur within a sentence. TextRank~\cite{mihalcea2004} is then applied to compute a salience score for each node. 
%Edge weights are the frequency of the two words co-occur. Then, a graph-based algorithm TextRank~\cite{mihalcea2004} that takes into account edge weights is applied to the graph to assign a score to each word. 
The second step generates and scores keyphrase candidates. A candidate keyphrase $k$ is scored by  summing the salience scores of the words it contains and normalising by its length (i.e., Eq.~\ref{eq:keyphrase_score}).
\begin{equation}\label{eq:keyphrase_score} 
\text{score}(k) =\frac{\sum_{w \in k}\text{TextRank}(w)} {\text{length}(k) +1}
\end{equation}

During the path selection process, we apply the same syntactic constraints as used for GraphPath, i.e., a candidate path must contain as least one top-15 topic words, one verb node, and satisfy the path length requirement. Finally, all candidate paths are re-ranked by normalising the total path weight over its compression length multiplied by the sum of the scores of the keyphrases it contains. The score of a sentence compression $c$ is finally given by:
% * <advaith@abdn.ac.uk> 2017-11-30T14:48:53.490Z:
% 
% The description doesn't make sense. Surely you want higher score if there are key phrases? Why are you penalising instead?
% 
% ^.
\begin{equation}\label{eq:sent_score} 
\text{score}(c) =\frac{\sum_{i,j \in \text{path}(c)} w_{(i,j)}} { \text{length}(c) \times \sum_{k \in c}  \text{score}(k)}
\end{equation}
where $w_{(i,j)}$ is the edge weight defined in Eq.~\ref{eq:edge_weight}.

\section{Experimental Setup}\label{sec:setup}

%We evaluated our topic labelling approach through an experiment where participants rated labels generated by our approaches as well as two baselines and two competitor systems.

%line 265-267 commented as non informative by reviewer.
\begin{table*}[tb] \centering \small\caption{Dataset statistics.}

%\begin{adjustbox}{width=0.71\textwidth,center=\textwidth} 

\begin{tabular}{lccccc} 
\hline
% \multirow{2}{*}{Dataset} & \multicolumn{3}{c}{Documents}             & \multirow{2}{*}{Vocab. size} \\ \cline{2-4}
%                          & \multicolumn{1}{l}{total num.} & \multicolumn{1}{l}{avg. \# of words} & \multicolumn{1}{l}{avg. \# of sentences} &                              \\ \hline
Dataset & total \# of doc.     & avg. \# of words & avg. \# of sentences &  Vocab. size                         \\ \hline

Kitchen             & 2,000                            & 25                                                       & 4                            & 2,450                           \\
Electronics              & 2,000                           & 25                                                   & 4                            & 2,317                           \\
IMDb            & 1,383                          & 118                                                     & 13                          & 14,337                           \\ \hline
\end{tabular}

%\end{adjustbox}

\label{table:dataset_stats}
\end{table*}
%\footnote{\url{https://www.cs.jhu.edu/~mdredze/datasets/sentiment/}}
%\footnote{\url{http://www.imdb.com/}}.
%\subsection{Setup}

\subsection{Dataset}
We evaluated the effectiveness of the sentence label selected by our approach on three real-world datasets.
Two datasets are publicly available, i.e., Amazon reviews for kitchen and electronic products, where each dataset contains 1000 positive and 1000 negative reviews\footnote{https://www.cs.jhu.edu/~mdredze/datasets/sentiment/}~\cite{blitzer2007biographies}. We have also collected a movie review dataset for \textit{Internal Affairs} and \textit{The Departed} from IMDb\footnote{http://www.imdb.com/}. 
These two movies virtually share the same storyline but with different casts and productions: \textit{Internal Affairs} is a 2002 Hong Kong movie, while \textit{The Departed} is a 2006 Hollywood movie. 

%\noindent\textbf{The Domesday dataset.}~~~In 1986, 900 years after William the Conqueror’s original Domesday Book, the BBC published the Domesday Project in an attempt to documenting everyday life in the United Kingdom. Over a million people contributed to record what they thought would be of interest in another 1000 years. The whole of the UK – including the Channel Islands and Isle of Man – was divided into 23,000 4x3km areas called Domesday Squares or “D-Blocks”. Schools and community groups surveyed over 108,000 square km of the UK and submitted more than 147,819 pages of text articles and 23,225 amateur photos, cataloguing what it was like to live, work and play in their community. Now, the Domesday dataset is available at a website called Domesday Reloaded run by the BBC.\footnote{\url{http://www.bbc.co.uk/history/domesday}}

\subsection{Implementation Details}

%\textbf{[Topic model setting for the experiments]} 
In the JST model implementation, we set the symmetric prior $\beta$=0.01~\cite{steyvers2007probabilistic}, the symmetric prior $\lambda= (0.05 \times L) / S$, where $L$ is the average of the document length, $S$ the is total number of sentiment labels, and the value of 0.05 on average allocates 5 percent of probability mass for mixing. The asymmetric prior $\alpha$ is learned directly from data using maximum-likelihood estimation~\cite{minka00estimate} and updated every 50 iterations during the Gibbs sampling procedure.  We empirically set the model with 2 sentiment labels and 10 topics for each sentiment label, resulting  20 sentiment-bearing topics in total for each of the three evaluation datasets.
%\footnote{\url{http://www.nltk.org/}}
In the preprocessing, we first performed automatic sentence segmentation\footnote{http://www.nltk.org/} on the experimental datasets in order to preserve the sentence structure information of each document, followed by parts of speech tagging on the datasets using the Stanford POS tagger~\cite{toutanova2003feature}.  Punctuation, numbers, and non-alphabet characters were then removed  and all  words were lowercased. Finally, we trained JST models with the preprocessed corpus following the standard procedures described in~\cite{Lin2011}, with the MPQA lexicon~\cite{wilson2005recognizing} being incorporated as  prior information for model learning.

%\begin{figure}[tb]
%	 \centering \label{fig:sent-results}
%    	\includegraphics[width = 1.2\textwidth]{sent-results.pdf}
% 	 \caption{(a) JST graphical model; (b) the generative process of JST.}  %% label for entire figure
%\end{figure}

\subsection{Baselines} \label{ssec:baselines}

\noindent\textit{Top probability}~~~The first baseline (\textbf{Top-prob}) is to select the sentence with highest topic relevance probability $p(\text{sent}|l,z)$ according to the JST model, as described in Section~\ref{sec:sentence_generate}. 
%As discussed  previously, this baseline does not guarantee selecting a sentence containing  a wide range of topic words. 

\noindent\textit{Centroid.}~~~The second baseline is a \textbf{centroid} based sentence label. For each sentiment-bearing topic, we first construct a sentence cluster consists of the top 150 most relevant  sentences for the given topic, ranked based on the topic relevance probability $p(\text{sent}|l,z)$. Next, for each of the sentence in the cluster, we compute the cosine similarity between the sentence and the remaining sentences in the cluster. Finally, the sentence with the highest cosine similarity score is picked as the topic label. This is a stronger baseline as it selects a sentence representative of multiple high probability sentences.

%We outline the steps for selecting a sentence label: (1) We use the \textbf{Top probability} method above to construct a list of 150 most relevant  sentences for the topic; (2) For each of these sentences,  we compute the cosine similarity to the whole list;  (3) We select the sentence with the highest cosine similarity value to the list as the topic label. This  is a stronger baseline as it selects a sentence representative of multiple high probability sentences.

%We use the cosine similarity computation for word distributions, previously used to compare two topics \cite{cano2014automatic}, to compare sentences.

\noindent\textit{Phrase/bigram baselines.}~~~In addition to the two sentential baselines, we also compare our approach to two systems which have been widely bench-marked in the topic labelling task, namely, \textbf{Mei07}~\cite{mei2007automatic}  and \textbf{Aletras14}~\cite{aletras2014}. Note that \textbf{Mei07} is  our re-implementation which generates bigram labels, while  \textbf{Aletras14} is the original implementation kindly provided by the authors which generates phrase labels. For both, we selected the top three bigrams/phrases for inclusion in the label.

\section{Experimental Results}\label{sec:results}

We evaluated our extractive topic labelling approach through an experiment where participants rated labels generated by our approaches as well as two baselines and two competitor systems in Section 5.1. 
%~\ref{ssec:extractive_results}.  
For the evaluation of our abstractive labelling approach, we added an additional Grammaticality measure as one of the evaluations for the label generated using the abstractive approach. The description of the evaluation is detailed in Section 5.2.
%~\ref{ssec:abstractive_results}.

\subsection{Evaluation of extractive sentence labelling}\label{ssec:extractive_results}

\addtolength{\tabcolsep}{-3pt}
\begin{table}[tb]
\caption{Human evaluation: Top-1 average quality ratings.}
%\textbf{[Fleiss Kappa = 0.52 evaluator agreement]}
%\vspace{-5pt}
\centering \small \label{tb:human-evaluation}
%\begin{adjustbox}{width=0.49\textwidth,left=\textwidth} 
\begin{tabular}{lccccc}
\hline
Domain  & Mei07& Aletras14& Top-prob& Centroid& Sent-label      \\ \hline
Kitchen  & 0.31       & 1.29     &  0.64 & 1.79   & 1.81  \\ 
Electronics  & 0.52      & 2.13      &  0.83 & 1.33   & 2.48  \\ 
IMDb & 0.36     & 1.73      &  0.69 & 1.67   & 2.58 \\ 
\hline
Average&0.39 &1.72 & 0.72&1.59 &2.29 \\

\end{tabular}
%\end{adjustbox}
\end{table}
\addtolength{\tabcolsep}{3pt}

% \begin{table*}[tb]
% \caption{Average label length across three datasets.}
% %\vspace{-5pt}
% \centering \small \label{tb:label-length}
% \begin{adjustbox}{width=1.0\textwidth,center=\textwidth} 
% \begin{tabular}{|l|c|c|c|c|c|c|c|}
% \hline
%   & Mei07           & Aletras14      &  Top-prob  & Centroid  & Sent-label &Path Graph &Keyphrase       \\ \hline
% Avg. label length  & 6     & 6.80     & 6.83 & 23.80  & 12.30&12.98  &7.80   \\ \hline
% \end{tabular}
% \end{adjustbox}
% \end{table*}

\begin{table}[tb] \centering \small 
\caption{Average label length in words across three datasets.} \label{tb:label-length}
%\vspace{-5pt}
%\begin{adjustbox}{width=0.49\textwidth,left=\textwidth} 
\begin{tabular}{ccccc}
\hline
  Mei07           & Aletras14      &  Top-prob  & Centroid  & Sent-label   \\ \hline
 6     & 6.80     & 6.83 & 23.80  & 12.70   \\
 \hline\end{tabular}
%\end{adjustbox}
\end{table}

%\subsubsection{Human Evaluation}

\noindent \textit{Method.}~~~For each sentiment-bearing topic, annotators were provided with the top 10 topic words with the highest marginal probability and  the labels generated by each system. They were then asked to judge the quality of each label on a 4-point Likert Scale: 

{\small
\begin{itemize}
    \item[3] Very good label, a perfect description of the topic;
    \item[2] Reasonable label, but does not completely capture the topic; 
    \item[1] Label is semantically related to the topic, but would not make a good topic label; and  
    \item[0] Label is completely inappropriate, and unrelated to the topic.
\end{itemize}
}
The most relevant review of the sentiment-bearing topic (calculated using $p(d|l,z)$, the probability of a document given a sentiment-bearing topic) was also provided to annotators so that they can have a better context for topic understanding. To ensure there was no bias introduced in this step,  we checked whether the most relevant review contains our sentence label. This only occurred in  7.5\%  of cases. For the remaining 92.5\%, sentence labels appeared in only the second to the fifth most relevant reviews, not shown to the user. This suggest that our experimental setup is not biased towards our system as the most relevant review rarely contains the most representative sentence label.

We recruited  30 Computing Science postgraduate students to participate in the evaluation task. Recalling that 20 sentiment-bearing topics were extracted for each dataset and that there are 5 systems to compare (i.e., our proposed approach and the four baselines) and 3 datasets, the total number of labels to rate is 300. To avoid the evaluation task being overly long, we broke the  evaluation task into 3 sessions carried out on different days, with  each session evaluating labels from one dataset only. Completing  one session took around 18 minutes for each participant on average. 

\noindent \textit{Quantitative Results.}~~~We report results using the Top-1 average rating~\cite{lau2011automatic}, which is the average human rating (based on the Likert Scale) assigned to the best candidate labels selected by each of the approaches, and the higher rating will indicate a better overall quality of a label. Table~\ref{tb:human-evaluation} shows the Top-1 average ratings for each approach (with our approach denoted as \textbf{Sent-label}). The sentence labels generated by our approach outperforms all baselines significantly  (two-tailed paired t-test; $p<0.001$ in all cases). Furthermore, as shown in Table~\ref{tb:label-length}, the label selected by our approach also keeps a good balance of information-richness and brevity, in contrast to the other two sentential baselines that favour either very short or very long sentences. To measure  inter-annotator agreement (IAA), we first calculated Spearman's $\rho$ between the ratings given by an annotator and the average ratings from all other annotators for the labels corresponding to the same topic. We then averaged the $\rho$ across annotators,  topics, and datasets, resulting in an average $\rho = 0.73$, a good IAA.  

We want to stress that these baselines are fair. To our knowledge, we are the first to study the problem of labelling sentiment-bearing topics, so there is no directly comparable system available which can generate sentence labelling for sentiment-bearing topics. In the past, top-$n$ topic terms have been commonly used for manually interpreting sentiment-bearing topics for a variety of sentiment-topic models (i.e., it's not just exclusively used for standard topic models), and that is why we propose an automatic labelling approach here. For the phrase label baselines,  they can capture a fair amount of sentiment information from sentiment-bearing topics based on the adjective/adverb phrases extracted. The sentence label baselines are also strong baselines as they both make use of the probability distributions from the JST model, with the centroid baseline further addressing the diversity issue.

\begin{table*}[!htb]
\centering \small  %\label{tb:label-example}
\caption{Labelling examples for sentiment-bearing topics.}
\label{tb:label-example}
%\resizebox{\textwidth}{!}{%
%\makebox[\linewidth][c]{%
%\hskip-1.5cm
%\hspace*{-0.6em}
%\begin{adjustbox}{width=1.1\textwidth,center=\textwidth} 
%\begin{tabular}{|c|l|p{0.39\linewidth}|p{0.52\linewidth}|}
\begin{tabular}{|c|l|p{0.38\linewidth}|p{0.38\linewidth}|}
\hline
 &  & \multicolumn{1}{c|}{\textbf{Positive Topics}} & \multicolumn{1}{c|}{\textbf{Negative Topics}} \\ \hline \hline
\multirow{6}{*}{\begin{sideways}Kitchen~~~~~~~~~~~~~~~~~~~\end{sideways}}  & \textbf{Topic words} & \textit{very good well really nice price quality set little happy} & \textit{money dont product disappointed back waste without very worth make} \\ \cline{2-4} 
 & \textbf{Mei07} & are very; a very; very good & the money; money on; you will \\ \cline{2-4} 
 & \textbf{Aletras14} & prices and quality; nice guy; coin grading & food waste; zero waste; waste your money \\ \cline{2-4} 
 & \textbf{Top-prob} & Well worth every penny. & Heres why I am sorely disappointed: \\ \cline{2-4} 
 & \textbf{Centroid} & I got this for the dogs and they seem to like it. &  I ordered the product and when I opened the box I found that I received only the bottom piece. \\ \cline{2-4} 
  %& \textbf{SumBasic} & Very good price too. &  Dont waste your money. \\ \cline{2-4} 
& \textbf{Sent-label} & Nice little toaster with good price. & Totally dissatisfy with this product and it's damn waste of money. \\ \hline \hline
\multirow{6}{*}{\begin{sideways}Electronics~~~~~~~~~~~~~~~~~~~~~~~~~~~~\end{sideways}} & \textbf{Topic words} & \textit{sound headphones music ear noise cord ipod pair quality head} & \textit{receiver slightly laser pointer green serious blocks kensington delormes beam} \\ \cline{2-4} 
 & \textbf{Mei07}& the sound; The sound; sound is & pointer is; of this; laser pointer \\ \cline{2-4} 
 & \textbf{Aletras14} & Noise-cancelling headphones; Previous Bose headphones; Bose headphones & Laser safety; Laser pointer; Laser diode \\ \cline{2-4} 
 & \textbf{Top-prob} & They have very good sound quality.& I made the mistake to buy the Kensington laser pointer before this one.\\ \cline{2-4}
 & \textbf{Centroid} & I just leave the mounts in place on the windshields of both cars and move the Nuvi and power cord to whichever car I am using at the moment.& I can maybe prefer a bit the ergonomy of the Kensington together with the very practical possibility to store the USB receiver inside the unit but this is probably just my taste and the hiro is quite comfortable either and lighter. \\ \cline{2-4} 
% & \textbf{SumBasic} & They have very good sound quality. &  I bought this one thinking that it would be really cool because it was green. \\ \cline{2-4} 
& \textbf{Sent-label} & The output of sound quality from an ipod is amazing. & The Kensington laser pointer is slightly  expensive. \\ \hline \hline
\multirow{6}{*}{\begin{sideways}IMDb~~~~~~~~~~~~~~~~~~~~~~~~~~~~\end{sideways}} & \textbf{Topic words} & \textit{without no want clever intelligent surely play offer effective puts} & \textit{language use vulgar blood used excessive dialog foul us realism} \\ \cline{2-4} 
 & \textbf{Mei07} & is one; that it; that there  & violence and; of violence; terms of \\ \cline{2-4} 
 & \textbf{Aletras14} & Flying Spaghetti Monster; Expelled: No Intelligence Allowed; Prayer& Vampire: The Masquerade ? Bloodlines; Monkeys in Chinese culture; Chavacano \\ \cline{2-4} 
 & \textbf{Top-prob} &But what would Lucifer be without his collaborators.& But the story flows well as does the violence. \\ \cline{2-4}
 & \textbf{Centroid} & The Departed is without doubt the best picture of the year and never fails to entertain. & It's far too graphically violent for teenagers, including images of heads being shot and spurting blood, limbs being broken, etc. \\ \cline{2-4} 
% & \textbf{SumBasic} & Jack Nicholson is just all right. &  I tried to get over the language. \\ \cline{2-4} 
 & \textbf{Sent-label} & If you want to see an intelligent, memorable thriller, watch The Departed. & The movie includes some overdone foul language and violent scenes. \\ \hline
\end{tabular}
%}
%}
%\end{adjustbox}
\end{table*}

%\noindent\textbf{Qualitative analysis.}~~~
\noindent \textit{Qualitative analysis.}~~~Table~\ref{tb:label-example} shows six sentiment-bearing topics for the Kitchen, Electronics and IMDb dataset extracted by JST. For each  topic, we also show the top-3 labels generated by Mei07 and Aletras14, two baselines (i.e., Centroid, Top-prob), and the sentence label generated by our approach (Sent-label). It can be seen from the table that the sentence labels generated by our approach generally  capture 
\noindent the opinions encoded in the sentiment-bearing topics quite well;  whereas in quite a few cases the labels generated by Mei07 and Aletras14 either only capture the thematic information of the topics   (e.g., \textit{Noise-cancelling headphones}) or merely sentiment information (e.g., \textit{very good}). 

Our labelling approach also shows better performance than the sentential baselines. For instance, both sentential baselines were unable to adequately interpret the negative sentiment-bearing topics in the Kitchen dataset. For example,  the Centroid baseline ``\textit{I ordered ... received only the bottom piece.}''  captures the thematic information (i.e., \textit{product}) but fails to capture the sentiment information (e.g., \textit{worth}). The Top-prob baseline captures negative sentiment information (i.e., \textit{disappointed}) but does not include any thematic information (e.g., \textit{money, product}). A user who read  the Top-prob baseline (i.e., ``\textit{Heres why I am sorely disappointed}'') is likely to have some confusions in understanding what causes the disappointment in the label, e.g., whether it is due to \textit{money} or \textit{product}. Our approach (i.e., Sent-label) also performs better in the Electronics dataset by giving more descriptive information about products. Take the positive sentiment-bearing topics for example,  the labels generated by Mei07 and Aletras14 capture  two different thematic aspects i.e., \textit{sound} and \textit{headphones}, respectively. For the Top-prob  label, it captures positive sentiment about \textit{sound quality}, whereas the Centroid baseline label seems less relevant to the topic. In contrast, our proposed approach shows that the topic conveys the opinion of  iPod sound quality being amazing.

To summarise, our experimental results show that our extractive sentence labelling approach outperforms four strong baselines and demonstrates the effectiveness of our sentence labels in facilitating topic understanding and interpretation.

\subsection{Evaluation of abstractive sentence labelling}\label{ssec:abstractive_results}

\begin{table*}[tb]
\centering \small  
\caption{Examples of extractive (Sent-label) and abstractive (PathGraph and Keyphrase) labels for sentiment-bearing topics.}
\label{tb:label-example-abstract}
\begin{tabular}{|l|p{0.40\linewidth}|p{0.35\linewidth}|}
\hline
& \multicolumn{1}{c|}{\textbf{Positive Topic}} & \multicolumn{1}{c|}{\textbf{Negative Topic}} \\ \hline \hline
 \textbf{Topic words} & \textit{pan heat cooking	oven	stick pans cook stainless food surface} & \textit{money dont product disappointed back waste without very worth make} \\ \hline
 \textbf{Sent-label} &I would have opted for the aluminum as that was what I had but my husband bought the stainless steel. & Totally dissatisfy with this product and it's damn waste of money. \\ \hline
 \textbf{PathGraph} &i would have opted for the aluminum as that was what i had but my husband bought the stainless steels great too of course but this is a worthy and earthy alternative.  & it was so not worth your money! \\ \hline
 \textbf{Keyphrase} & it is a good quality stainless and that is important. &very disappointed in product. \\ \hline 
\end{tabular}
\end{table*}

\addtolength{\tabcolsep}{+3pt}

\begin{table}[tb]
\centering \small
\caption{Human evaluation: Grammaticality ratings, expressed on a scale of 0 to 2. The average ratings (Avg.) are also reported.}%and  the inter-annotator agreement ($k$) are also reported.}
\label{human_grammaticality}
%\begin{adjustbox}{width=0.45\textwidth,left=\textwidth} 
\begin{tabular}{lcccccccccc}
\hline
 & \multicolumn{3}{c}{Grammaticality Rating} &  &  \\ \cline{2-4}
System & 0 & 1 & 2 & Avg.  \\ \hline
PathGraph  & 7\% & 45\%& 47\% & 1.40  \\ 
Keyphrase  & 7\% & 35\% & 57\%& 1.49 \\ 
Sent-label & 0\% & 19\% & 79\% & 1.80  \\ \hline
 &  &  &  &  & 
\end{tabular}
%\end{adjustbox}
\end{table}

\begin{table}[tb]
\caption{Average label length in words of extractive and abstractive labels across three datasets.}
%\vspace{-5pt}
\centering \small \label{tb:abs-label-length}
%\begin{adjustbox}{width=0.45\textwidth,left=\textwidth} 
\begin{tabular}{ccc}
\hline
Sent-label &PathGraph &Keyphrase       \\ \hline
12.70&13.87 & 9.90    \\ \hline
\end{tabular}
%\end{adjustbox}
\end{table}

\begin{table}[tb]
\small
\caption{Human evaluation: Informativity ratings for extractive and abstractive labels based on Top-1 average rating.}
\centering \small \label{tb:human-evaluation_2}
%\begin{adjustbox}{width=0.45\textwidth,center=\textwidth} 
%\begin{adjustbox}{width=0.35\textheight,left=\textwidth} 
\begin{tabular}{lccc}
\hline
Domain  &   Sent-label &PathGraph &Keyphrase     \\ \hline
Kitchen  &     2.00 &2.10 &2.47 \\ 
Electronics  & 1.85 &1.82&2.10           \\ 
IMDb &     2.14 &1.92  &2.61       \\ 
\hline
Average& 1.90 &1.94&2.39 \\

\end{tabular}
%\end{adjustbox}
\end{table}

\noindent\textit{Method.}~~~To evaluate the effectiveness of the abstractive labels, we compared the compressions generated by GraphPath and Keyphrase against the best performing extractive label (Sent-label) in terms of grammaticality and informativity. In the evaluation,  
%An example of almost grammatical sentence is shown in Table~\ref{tb:label-example2}. 
ten raters who are native speakers were presented with a list of top-10 topic words along with one label generated from the following labelling approaches (i.e., Sent-label, PathGraph  and Keyphrase) in a random order. For each label, raters were asked to rate the label grammaticality and the label coverage by using the topic words as guidance. In addition, raters were explicitly asked to ignore lack of capitalisation while evaluating grammaticality. 
The grammaticality of the labels are rated on a 3-points Likert scale: 
\begin{itemize}
    \item[2] \textit{good sentence} , if the label is a complete grammatical sentence; 
    \item[1] \textit{almost}, if the label can be understand but requires minor editing, e.g. one mistake in articles; \item[0]\textit{ungrammatical}, if it is none of the above.
\end{itemize}
For informativity, labels were rated using the same 4-points Likert scale as specified in Section~5.1.
%~\ref{ssec:extractive_results}.

\noindent\textit{Results.}~~~ Table~\ref{human_grammaticality} presents the average grammaticality scores of the sentence labels generated by three different approaches. Unsurprisingly, Sent-label achieves the highest grammaticality score as its labels are the original sentences extracted from the corpus. Although the abstractive methods have lower grammaticality scores, almost 60\% of the sentences generated by the Keyphrase algorithm are  perfectly grammatical. When comparing the abstractive methods, it is observed that the grammaticality score of Keyphrase is slightly higher than that of PathGraph. This is probably due to the fact that  Keyphrase compression generates shorter sentence labels than PathGraph (cf. Table~\ref{tb:abs-label-length}). Table~\ref{tb:human-evaluation_2} shows the human evaluation results for informativity based on Top-1 average rating. We  observe a large improvement in informativity for the Keyphrase algorithm, representing an absolute  increase of 0.45 over the both Sent-label and PathGraph. It should also be noted that the average length for Keyphrase is  shorter than the average length for both Sent-label and PathGraph, meaning that Keyphrase can better generate labels with high information coverage of the topic.

Table~\ref{tb:label-example-abstract} shows some examples of extractive (i.e., Sent-label) and abstractive (i.e., PathGraph and Keyphrase) labelling for two sentiment-bearing topics extracted from the Kitchen dataset. 

To summarise, our experimental results show that the Keyphrase algorithm outperforms the PathGraph algorithm in terms of both grammaticality and informativity. In addition, the abstractive labels generated based on the compression algorithms are more informative than the extractive label, with some cost of reduced grammaticality.

\section{Conclusion and Future Work}\label{sec:conclusions}

This paper tackles the problem of automatically labelling sentiment-bearing topics with descriptive sentence labels. We propose two approaches to the problem, one extractive and the other abstractive. Both approaches rely on a novel mechanism to automatically learn the relevance of each sentence in a corpus to sentiment-bearing topics extracted from that corpus. 
%An advantage of the proposed fusion approach is that it is requires only a Part-Of-Speech (POS) tagger and a list of stopwords to generate abstractive sentence labels. 
To our knowledge, we are the first to study the problem of labelling sentiment-bearing topics. Our  experimental results on three real-world datasets show that both the extractive and abstractive approaches  outperform four strong baselines in terms of facilitating topic understanding and interpretation. In addition, when comparing extractive and abstractive labels,  abstractive labels are able to provide more topic information coverage despite being shorter, as they can synthesise information from different sentences needed for sentiment-bearing topic interpretations. It should also be noted  that our approach does not have any specific dependencies on the JST model, and thus it is general enough to be directly applied to any other sentiment topic model variants which generate multinomial topics as output. 
%which can facilitate the task of interpreting the opinions encoded in sentiment-bearing topics.

In the future, we would like to extend our work for opinion summarisation. One natural way of achieving this is to summarise documents through the propagation of the document-topic and topic-sentence associations learned from our framework. It is also possible to improve the relevance measure between sentiment-bearing topics and sentences by leveraging external knowledge~\cite{lin2015sherlock,mao2018word} in addition to textual features. 

%\textbf{[TODO: say something about summarisaiton as future work.]}
%In the future, we would like to extend this work and adapt it to the opinion summarisation task.  We would like to find out how successful this work in solving opinion summarisation task and explore the connection between sentiment-bearing topics and opinion summarisation task performance. Our experiment shows that the descriptive sentence labels can be exploited to provide a better label to interpret the sentiment-bearing topics. 

\Acknowledgements {This work is supported by the award made by the UK Engineering and Physical Sciences Research Council (Grant number: EP/P005810/1).}
\bibliographystyle{unsrt_Modhardy}
\bibliography{bib/fcs}

\end{document}